# Three studies
# of grammar-based surface parsing
# of unrestricted English text

Atro Voutilainen

May 5, 1994

# Abstract

Three studies of grammar-based surface parsing of unrestricted English text

Voutilainen, Atro Tapio
University of Helsinki, SF

The dissertation addresses the design of parsing grammars for automatic surface-syntactic analysis of unconstrained English text. It consists of a summary and three articles.

*Morphological disambiguation* documents a grammar for morphological (or part-of-speech) disambiguation of English, done within the Constraint Grammar framework proposed by Fred Karlsson. The disambiguator seeks to discard those of the alternative morphological analyses proposed by the lexical analyser that are contextually illegitimate. The 1,100 constraints express some 23 general, essentially syntactic statements as restrictions on the linear order of morphological tags. The error rate of the morphological disambiguator is about ten times smaller than that of another state-of-the-art probabilistic disambiguator, given that both are allowed to leave some of the hardest ambiguities unresolved. This accuracy suggests the viability of the grammar-based approach to natural language parsing, thus also contributing to the more general debate concerning the viability of probabilistic vs. linguistic techniques.

*Experiments with heuristics* addresses the question of how to resolve those ambiguities that survive the morphological disambiguator. Two approaches are presented and empirically evaluated: (i) heuristic disambiguation constraints and (ii) techniques for learning from the fully disambiguated part of the corpus and then applying this information to resolving remaining ambiguities.

*Designing a parsing grammar* starts with a critical evaluation of the merits and problems of Constraint Grammar. The reductionistic grammar-based style of analysis at a structurally motivated level of description is found to be viable. The negative criticisms mainly concern the sequentiality/modularity of the framework and the relative unexpressiveness of the parsing scheme used in the English Constraint Grammar description. Recent work is documented within the Finite-state Intersection Grammar framework, proposed by Kimmo Koskenniemi. A new expressive and resolvable dependency-oriented functional grammatical representation is outlined, and substantial examples are presented from an emerging comprehensive finite-state grammar of English.

# Preface

The work reported in this dissertation was carried out at the Research Unit for Computational Linguistics at the University of Helsinki, 1989–1993. The work was funded by the Academy of Finland and the Technology Development Centre of Finland (TEKES).

I wish to thank Dr. Roger Garside, Prof. Geoffrey Leech and the UCREL team at Lancaster University for their permission to use the CLAWS1 tagger, and Dr. Ken Church and the AT&T Bell Labs for their permission to use the *Parts of speech* tagger in my work. I would also like to thank Prof. Eva Ejerhed from Umeå University and Doc. Gunnel Källgren from Stockholm University for their insightful previews of the prefinal version of this dissertation.

I owe a debt of gratitude to my instructors, Prof. Fred Karlsson and Prof. Kimmo Koskenniemi, for invaluable guidance, criticisms and numerous discussions.

I am most grateful to Arto Anttila, Juha Heikkilä, Pasi Tapanainen, Jussi Piitulainen, Krister Lindén, Timo Järvinen, Kari Pitkänen, Tapani Kelomäki and Martti Nyman for their comments on various parts of the dissertation. I am grateful to Dr. Andrew Chesterman for improving the English of the Summary.

Finally, I would like to thank Ms. Sanna Räsänen for her encouragement, and my parents Liisa and Väinö for their support over the years.

Helsinki, February 1994          A.V.

ii

# Contents

# Chapter 1

# Introduction

This paper is a summary of and orientation to the following three papers by the present author:

1. Morphological disambiguation. In Karlsson, Voutilainen, Heikkilä and Anttila (eds.) (1994). *Constraint Grammar: a Language-Independent System for Parsing Unrestricted Text.* Berlin and New York: Mouton de Gruyter.

2. Experiments with heuristics. In Karlsson, Voutilainen, Heikkilä and Anttila (eds.) (1994). *Constraint Grammar: a Language-Independent System for Parsing Unrestricted Text.* Berlin and New York: Mouton de Gruyter.

3. *Designing a parsing grammar.* Publications of the Department of General Linguistics, No. 22, University of Helsinki. 1994.

Before starting with the summary, some background facts are in order. From 1989 to 1992, four researchers – Fred Karlsson, Arto Anttila, Juha Heikkilä and myself – from the Research Unit for Computational Linguistics at the University of Helsinki participated in the ESPRIT II project No. 2083 SIMPR (*Structured Information Management: Processing and Retrieval*). Our task was to make an operational parser for running English text mainly for information retrieval purposes.

The parsing framework, known as Constraint Grammar, was originally proposed by Karlsson [1990], and he also wrote a LISP version of the Constraint Grammar Parser. Let us outline a few characteristics of Constraint Grammar:

- Morphological and syntactic descriptions are encoded with tags rather than e.g. with phrase structure bracketing. The syntactic descriptions are very shallow; in this way many structurally unresolvable ambiguities are left implicit. Each word is flanked with a syntactic function tag that imposes an underspecific functional dependency-oriented description on the sentence. No special status is granted to phrases or clauses. A simple example is in order.

```
("<*the>"
    ("the"  <Def> DET CENTRAL ART SG/PL (@DN>)))
("<fat>"
    ("fat" A ABS (@AN>)))
("<butcher's>"
    ("butcher" N GEN SG (@GN>)))
("<wife>"
    ("wife" N NOM SG (@SUBJ)))
("<ate>"
    ("eat" <SVO> <SV> V PAST VFIN (@+FMAINV)))
("<an>"
    ("an" <Indef> DET CENTRAL ART SG (@DN>)))
("<apple>"
    ("apple" N NOM SG (@NN>)))
("<pie>"
    ("pie" N NOM SG (@OBJ)))
("<$.>")
```

In a morphologically and syntactically analysed representation, each word is flanked with a base form, morphological tags indicating part of speech, inflection, derivation etc., as well as with syntactic function tags. For instance, $@AN >$ indicates that *fat* is an adjectival premodifier of the next nominal head in the right-hand context or of an intervening premodifier; *wife* is a subject; *ate* is a past tense verb functioning as a finite main verb ($@ + FMAINV$) in the clause.

- The analysis of running text is a central concern. It is considered preferable to succeed in assigning a reliable shallow analysis rather than fail in imposing a highly structured, hierarchical analysis.

- Grammatical analysis is based on disambiguation. First, for each input word-form, all possible morphological and syntactic readings are provided as alternatives by a Koskenniemi-style morphological analyser and by simple mapping operations. For instance, here is the morphologically analysed representation of the sentence *That round table might collapse*:

```
("<*that>"
    ("that" <*> <**CLB> CS (@CS))
    ("that" <*> DET CENTRAL DEM SG (@DN>))
    ("that" <*> ADV AD-A> (@AD-A>))
    ("that" <*> PRON DEM SG)
    ("that" <*> <NonMod> <**CLB> <Rel> PRON SG/PL))
("<round>"
    ("round" <SVO> <SV> V SUBJUNCTIVE VFIN (@+FMAINV))
```

```
      ("round" <SVO> <SV> V IMP VFIN (@+FMAINV))
      ("round" <SVO> <SV> V INF)
      ("round" <SVO> <SV> V PRES -SG3 VFIN (@+FMAINV))
      ("round" PREP)
      ("round" N NOM SG)
      ("round" A ABS)
      ("round" ADV ADVL (@ADVL)))
("<table>"
      ("table" N NOM SG)
      ("table" <SVO> V SUBJUNCTIVE VFIN (@+FMAINV))
      ("table" <SVO> V IMP VFIN (@+FMAINV))
      ("table" <SVO> V INF)
      ("table" <SVO> V PRES -SG3 VFIN (@+FMAINV)))
("<might>"
      ("might" <-Indef> N NOM SG)
      ("might" V AUXMOD VFIN (@+FAUXV)))
("<collapse>"
      ("collapse" N NOM SG)
      ("collapse" <SV> <SVO> V SUBJUNCTIVE VFIN (@+FMAINV))
      ("collapse" <SV> <SVO> V IMP VFIN (@+FMAINV))
      ("collapse" <SV> <SVO> V INF)
      ("collapse" <SV> <SVO> V PRES -SG3 VFIN (@+FMAINV)))
("<$.>")
```

The second main module is syntactic analysis. Syntactic (including part-of-speech) analysis means discarding illegitimate alternatives on the basis of structural information in the context; what survives the reductionistic parser is the preferred syntactic analysis. The correct analysis is already in the parser's input; the task of the parser is simply to find the correct parse by 'sculpting' off the excess, that is, the illegitimate alternatives. After morphological (or part-of-speech) disambiguation, but before the introduction of syntactic ambiguities, the above sentence would look like this:

```
("<*that>"
      ("that" <*> DET CENTRAL DEM SG (@DN>)))
("<round>"
      ("round" A ABS))
("<table>"
      ("table" N NOM SG))
("<might>"
      ("might" V AUXMOD VFIN (@+FAUXV)))
("<collapse>"
      ("collapse" <SV> <SVO> V INF))
("<$.>")
```

- The parsing descriptions (linguistic rule components, especially the lexicon and the parsing grammar) are based on linguistic generalisations rather than probabilities automatically generated from tagged text corpora. Grammatical generalisations are expressed as linear order constraints on distributional categories. In Constraint Grammar, these constraints express partial facts in a negative, reductionistic fashion. For instance, a constraint for part-of-speech disambiguation might state a partial generalisation about the form of a noun phrase in a negative fashion by disallowing a finite verb reading that directly follows an unambiguous determiner.

- Parsing is modular and sequential, e.g. morphological or part-of-speech disambiguation precedes syntactic analysis. Also the grammar is built of separate subgrammars for

  1. morphological disambiguation,
  2. recognition of sentence-internal clause boundaries, and
  3. syntactic analysis.

The English Constraint Grammar description ENGCG was written by Anttila, Heikkilä and myself. My task was to write the morphological description and the lexicon with Heikkilä, and I also wrote the grammar for morphological disambiguation. Anttila wrote the grammar for syntactic functions. This dissertation is primarily concerned with grammar; the lexical description is documented elsewhere [Voutilainen and Heikkilä, 1994a; Heikkilä, 1994a; Heikkilä, 1994b].

<p style="text-align:center">***</p>

The chapter *Morphological disambiguation* documents the grammar for morphological disambiguation by showing on the one hand how linguistic generalisations can be expressed as constraints, and, on the other, by showing what the underlying grammatical generalisations are. The accuracy of the description is empirically tested and compared to two state-of-the-art probabilistic systems.

The grammar-based disambiguator leaves a word ambiguous rather than risks an uninformed guess. The chapter *Experiments with heuristics* presents some solutions to the resolution of these remaining ambiguities.

The third study, *Designing a parsing grammar*, takes a critical view on some features in the Constraint Grammar framework and in the ENGCG description. Arguments are provided for a parallel and more expressive representation of grammatical relations in a surface-oriented parsing grammar, and emerging work within the Finite-state Intersection Grammar framework, originally proposed by Koskenniemi [1990], is documented. This study ends with an outline of a new, more expressive grammatical representation and substantial examples from an emerging finite-state grammar of English.

# Chapter 2

# Summary

This section outlines each of the three papers that comprise my dissertation. The logical structure of this summary reflects that of the articles.[1]

## 2.1 Morphological disambiguation

This long chapter in [Karlsson, Voutilainen, Heikkilä and Anttila (eds.), 1994] documents the English disambiguation grammar written within the Constraint Grammar framework proposed by Karlsson [1990].

*METHODOLOGICAL GUIDELINES*

This section is a Constraint Grammarian's manual. Practical guidelines are provided for writing and evaluating a disambiguation grammar.

- Morphological ambiguity in English is the consequence of employing descriptors in a morphological description that are not entirely predictable from the form of the word itself. On an average, the ENGTWOL two-level morphological analyser of English provides each word-form token in running text with approximately two alternative morphological readings.

- The task of the morphological disambiguator is to discard all and only the contextually illegitimate alternative morphological readings. The disambiguator employs an unordered set of linguistic constraints on the linear order of ambiguity-forming morphological readings.

- Evaluation metrics are defined:

---

[1]For the sake of clarity, also some additional comments are included.

- *Recall*: The ratio 'received appropriate readings / intended appropriate readings'.

- *Precision*: The ratio 'received appropriate readings / all received readings'.

A recall of less than 100% indicates that some legitimate analyses were lost, and a precision of less than 100% indicates that some illegitimate analyses survived the disambiguator.

- Specifying the grammatical representation.[2]

  - The 'correct' or appropriate analysis of linguistic utterances is a theoretical matter. Specifying a grammatical representation involves making

    > ... a statement of just what range of structures and categories will be used ... and which among various analyses in terms of those structures and categories will be regarded as correct for any debatable construction ... [Sampson, 1987b, p. 82]

    In practice, this means creating a representative annotated and carefully documented grammar definition corpus.

  - At least at a low level of analysis, it is possible to reach a near-100% consensus on what is the appropriate analysis in the analysis of running text. Two experiments on this controversial issue are reported.

- Writing and testing a disambiguation grammar.

  - Because there is no rule order in the grammar, the effects of each constraint are easy to observe. Ideally, each constraint is true, so the constraints can be applied in any order during parsing without changing the final analysis.

  - Ambiguity that survives the current disambiguator usually means that the distribution of some ambiguity-forming feature is not sufficiently accounted for in the grammar, i.e., new restrictions on the linear order of tags should be imposed. Constraints can refer to words and tags at fixed or relative word positions. This flexibility makes it possible to state grammatical generalisations in enough detail to make for a relatively accurate parsing grammar.

  - Writing a disambiguation grammar for all varieties of the language at once is impracticable. The proper object language should be specified in sufficient detail.

---

[2]Actually, the term 'parsing scheme' is used in the chapter itself, but since then I have come to prefer the term 'grammatical representation' because of its lack of connotation to the parsing aspect.

- The current English disambiguator avoids making risky predictions; rather, an ambiguity is left pending. Unresolved ambiguities in the present system can be classified into three categories:

  1. Structurally genuine global ambiguities. The knowledge-based resolution of these ambiguities usually presupposes semantic or world knowledge.

  2. Ambiguities due to some undesirable property of the Constraint Grammar formalism. For instance, the representation of ambiguities only at the word level compromises the disambiguator's accuracy.

  3. Ambiguities due to the unexpressiveness of the available grammatical representation. Morphological or part-of-speech disambiguation appears to be essentially syntactic. In the absence of the syntactic representation (syntactic functions, representation of phrase and clause level categories), it is sometimes prohibitively difficult to express the relevant generalisations as constraints.

  Relatively little can be done about genuine global ambiguities, but the other two problems should be possible to address by improving the design of the parsing formalism and the grammatical representation.

- The role of ill-formed input.

  - All input receives some analysis in Constraint Grammar parsing. The parsing grammar itself is a partial and roundabout description of what is considered (relatively) acceptable, i.e. the grammar disallows many ill-formed analyses. However, the application of the grammar can be constrained: the parser will not consult the grammar in the case of unambiguous analyses.

  - The centrality, or frequency, of ill-formed input in texts in the object language depends on what is considered ill-formed. The present grammar usually accounts for what occurs relatively frequently in running text, irrespective of stylistic etc. judgements, so what is considered genuinely ill-formed does not often occur in input texts, i.e. the consequences of ill-formed input do not appear particularly significant in the ENGCG description.

- Current status of the grammar.

  - The present grammar of 1,100 constraints is based on descriptive grammars and studies of various corpora.

- Of all words, 93–97% become unambiguous. At least 99.7% of all words retain the contextually most appropriate morphological reading after morphological disambiguation.[3]

- There is also an optionally applicable heuristic grammar of 200 constraints that resolves about half of the remaining ambiguities 96–97% reliably. In other words, after the application of the heuristic constraints, the recall of the entire system from morphological analysis through heuristic disambiguation is more than 99.5%, with a precision of about 96–98%.[4]

- Comparing ENGCG to state-of-the-art systems in terms of accuracy is not quite straightforward e.g. because, unlike most of its competitors, ENGCG does not even attempt to resolve all ambiguities, but rather leaves an uncertain ambiguity pending.[5] Luckily, there is a probabilistic state-of-the-art tagger whose recall is reported also for output with pending ambiguities, namely de Marcken's tagger [1990]. With exactly one analysis per word in its output, de Marcken's tagger has a recall of approximately 96%. With an average 1.04 alternative analyses per word in its output, de Marcken's tagger has a recall of approximately 97.6%. In the test reported in the chapter, ENGCG also left an average 1.04 alternative analyses per word in its output, with a recall of 99.8%. In other words, the ENGCG morphological disambiguator seems to have an error rate (number of discarded legitimate readings) of at least ten times smaller than that of a competing probabilistic state-of-the-art tagger.[6]

## A DISAMBIGUATION GRAMMAR OF ENGLISH

The grammar for morphological disambiguation is examined from two perspectives. Firstly, the analysis of a sample sentence is examined in detail to get an understanding of the rule formalism as a grammarian's tool and of the nature of the actual constraints. The second part of this section presents the generalisations underlying the disambiguation grammar. It is also shown how to express these 23 underlying generalisations as constraints.

[3]Note that these figures also include errors due to the ENGTWOL lexicon and 'morphological heuristics', a rule-based module that assigns ENGTWOL-style analyses to those words not represented in ENGTWOL itself.

[4]Currently, ENGCG contains no module that disambiguates the remaining 2–4%. If a blind guessing module were used, the overall recall and precision of the entire system from morphology through the guessing module, i.e. with no ambiguity in the output, would be 98% or a little more. – Using a probabilistic tagger such as CLAWS1, which, as suggested in *Morphological disambiguation*, may be able to predict with about 80% reliability about those ambiguities left pending by the ENGCG morphological disambiguator [Marshall, 1983; Garside, Leech and Sampson (eds.), 1987] instead of a blind guessing module would probably result in a recall and precision of about 99%.

[5]There certainly are other problems for comparisons between different systems as well, e.g. differences in tag sets.

[6]Also errors due to morphological analysis, incl. morphological heuristics, are included in these figures.

- Sample analysis. The disambiguation of a lengthy text sentence is examined. Observations:

  - The effect of the disambiguator was reasonably good. Almost all ambiguities were resolved, and no correct readings were discarded.

  - The constraints used in disambiguating the sentence seem to be very roundabout and partial expressions of higher-level syntactic relations. The roundaboutness is mainly due to the unavailability of the syntactic and clause boundary representation.

  - Most constraints are overtly negative, and often somewhat lacking in transparency: rather than positively express distributions for the described categories, they explicitly disallow illegitimate sequences of tags or words.

  - Some constraints are based on somewhat simplified linguistic generalisations in order to maximise both recall and precision.

  - Overall, the rule formalism seems to be at its best in the description of relatively local phenomena, e.g. the form of the simple noun phrase. Generalisations involving phrases and clauses are more difficult to express. Some ambiguities remain unresolved because of this.

- Constraint typology. The grammatical generalisations underlying the present 1,100 disambiguation constraints are presented. Certain phenomena such as ellipsis are not accounted for in the parsing grammar, i.e. some of the underlying generalisations are not entirely true.[7] It is also shown how these general statements can be expressed as constraints. The grammar partially expresses 23 generalisations:

  - Nominal phrase

    * *A postmodifier is preceded by its head or another, potentially coordinated postmodifier. In between, premodifiers of adjectives, adverbs and quantifiers are allowed.*

    * *An AD-A premodifies an adjective, adverb or quantifier.*

    * *Determiners and premodifiers are followed by a nominal head. In between, only certain (potentially coordinated) determiners and premodifiers are legitimate.*

    * *Nonmodifiable pronouns (those with the feature <NonMod> ) and proper nouns in general do not have determiners or premodifiers.*

---

[7]Some less frequent constructions are ignored in order to maximise the overall accuracy of the disambiguator.

* *A predeterminer is immediately followed by a central determiner, postdeterminer, premodifier, (with AD-As of its own), nominal head, or a coordinated predeterminer.*
* *Central determiners and postdeterminers are immediately followed by a postdeterminer, a premodifier (with AD-As of its own), a nominal head, or a coordinated determiner.*

– Verb chain

* *In declarative clauses, an auxiliary is followed by a main verb. In between, only adverbials and other (potentially coordinated) auxiliaries may occur.*
* *To the right of an infinitive marker, there is an infinitive.*
* *The infinitive marker* to, in=order=to *etc. is followed by an infinitive. In between, only an adverb can occur.*
* *To the left of an infinitive, there is an infinitive marker, a modal auxiliary, a verb taking an infinitive, or a coordinated infinitive.*
* *To the right of an auxiliary, there is a main verb.*

– Clause, sentence

* *A subordinator is in a finite clause, to the left of the finite predicate.*
* *Imperatives occur in subjectless main clauses, to the left of all clause-level nominal constituents.*
* *Each simplex finite clause contains exactly one finite verb.*
* *Predicate complements mainly occur towards the end of the clause.*
* *In finite clauses in the indicative, interrogative or subjunctive mood, there is a nominal or an adverbial with a nominal function to the left of the main predicate verb other than 'do', 'be' or 'have'.*
* *Subjunctives occur in clauses with* that *or* lest *as subordinating conjunction.*
* *A sentence contains at least one (potentially coordinated) main clause.*

– Agreement

* *An accusative is preceded by a main verb or a preposition or a coordinated accusative.*
* *A verb in the present tense agrees with the subject in number.*
* *Determiners agree in number with their heads.*

– Coordination

* *Only likes coordinate.*

– Complementation

* *A preposition is immediately followed by a coordinated preposition or a noun phrase acting as a complement.*

This final main section reports on a comparison between three taggers. The application of the ENGCG morphological disambiguator as well as of two high-performance probabilistic systems (CLAWS1 by the UCREL team [Marshall, 1983; Garside, Leech and Sampson (eds.), 1987] and *Parts of speech* by Church [1988]) on a test corpus is reported.

- A short introduction to CLAWS1 and *Parts of speech* is given. Both systems are based on probabilistic techniques, and their accuracy is of state-of-the-art quality (95–97% accuracy).

- An explicit comparison between different taggers presupposes a metric for the resolvability of the different grammatical representations. No such metric is available, therefore this comparison remains somewhat informal.

- The grammatical representations of the systems are compared. The substantial differences are not very considerable. The ENGCG grammatical representation appears to be the most detailed of the three, while the representation used in *Parts of speech* seems to be the most ascetic.

- Five texts totalling 2,167 words were used. These texts had not been used in the development of ENGCG before the test.

- The results are in agreement with previous reports (see Figure 2.1).

Figure 2.1: Performance of three taggers on a 2,167-word test corpus.

|  | *Recall* | *Precision* |
|---|---|---|
| CLAWS1 | 96.95% | 96.95% |
| ENGCG | 99.77% | 95.54% |
| *Parts of speech* | 96.21% | 96.21% |

- The probabilistic systems made a misprediction in 3–4% of all words. ENGCG, on the other hand, made a misprediction in only 0.23% of all words, but left more than 4% of all words ambiguous.

## 2.2  *Experiments with heuristics*

This chapter in [Karlsson, Voutilainen, Heikkilä and Anttila (eds.), 1994] documents three heuristic techniques, two of which have been fully incorporated in the ENGCG analyser.

*MORPHOLOGICAL HEURISTICS*

This section documents a mechanism for assigning ENGTWOL-style analyses to words not represented in the ENGTWOL description itself.

- The mechanism has the following properties:

    - Rules are ordered. This facilitates the expression of default rules.

    - The rules can refer to the form and context of the unrecognised word.

    - One or more alternative ENGTWOL-style readings can be assigned.

- The English description is outlined:

    - The classification resembles the ENGTWOL description, but it is less distinctive in the description of nominals (no explicit difference is made common nouns, proper nouns and abbreviations).

    - Most of the rules refer entirely to the form of the word, mainly to the beginning and ending.

    - Most unrecognised words are nouns and abbreviations. The general strategy is to recognise other categories – mainly adjectives and verbs – on the basis of word form, and describe the residual as nouns.

    - Each analysis is flagged with a special symbol to indicate its origin.

    - No attempt is made at spell-checking. The input text should be spell-checked beforehand.

- A performance test was made. A 16,000-word text was analysed with the ENGTWOL morphological analyser. About 3% of all word-form tokens were not recognised. Morphological heuristics were applied on these remaining 476 word-forms; 99.5% of these heuristic predictions were correct. The effect of this error rate on the overall performance of the ENGCG parser is almost negligible.

The rest of *Experiments with heuristics* is concerned with the resolution of those ambiguities that survive the disambiguation grammar documented in *Morphological disambiguation*. It is suggested in *Morphological disambiguation* that a probabilistic system could resolve these pending ambiguities with a little above 80% reliability. The remaining sections in the chapter on heuristics investigate the possibility of developing techniques for resolving at least some of these ambiguities more reliably than the expected 80%.

## HEURISTIC DISAMBIGUATION CONSTRAINTS

This section documents a heuristic grammar for resolving those ambiguities that survive the grammar-based constraints.

- Formally, heuristic disambiguation constraints are like grammar-based ones.

- The English heuristic description is outlined:

  - The heuristic grammar contains some 200 constraints.

  - The constraints are based on simplified linguistic generalisations. Less frequent distributions of the ambiguity-forming categories are ignored.

  - The following three heuristic strategies are illustrated:

    1. *Prefer common word orders.*
    2. *Prefer shallow analyses.*
    3. *Prefer common form–function assignments.*

  - The description is robust but not quite as mature as the collection of 1,100 constraints.

- A performance test is reported:

  - A collection of texts new to the system, 30,000 words in all, is used as test material.

  - The grammar-based disambiguator left 5% of all words ambiguous.

  - The heuristic constraints resolved about 50% of the remaining ambiguities. Of all these heuristic predictions, 96% were correct, i.e. considerably better than the 80% 'benchmark'.

  - After the application of both grammar-based and heuristic constraints, 96–98% of all words are morphologically unambiguous, with at least 99.5% of all words retaining the correct morphological analysis.

## TEXT-BASED DISAMBIGUATION

A common practice in statistical language analysis is to generate statistics from a tagged, heterogeneous corpus like Brown and LOB, and apply these statistics in the analysis of new texts. If the new text is different from the statistical model of the source corpus, the analyser is likely to perform less satisfactorily.

A new possibility is to use the analysed corpus itself both as a source of generalisations and as an object of the analyser based on these generalisations. The section *Text-based disambiguation* investigates the possibility of making generalisations from that part of the text fully disambiguated by the reliable grammar-based disambiguator, and of applying these generalisations to the resolution of the pending ambiguities. Two empirical case studies are reported.

- Text-based lexical probabilities.

  - What may appear as a relatively unbiased part-of-speech homograph in a heterogeneous corpus may be used in a far more uniform fashion in a particular text. Lexical probabilities based on heterogeneous corpora may therefore be unreliable in the analysis of a single text.

  - A study of noun–verb ambiguities is documented.

    * Statistics on part-of-speech distribution are generated from the fully disambiguated part of the text, and the proper analysis is selected according to the majority principle. A predominance threshold is used: a part of speech is selected only if it occurs at least twice as often as its competitors.

    * A test with another 29,000-word text was made. After grammar-based ENGCG disambiguation, some 6% of all words remained ambiguous. Applying text-based lexical probabilities to noun–verb ambiguous words fully disambiguated 36% of the remaining ambiguous words. Of these predictions, 95.5% were correct, i.e. again a definite improvement over the assumed 80% 'benchmark'.

- Disambiguating with collocations.

  - Sometimes a word that occurs in a collocation remains ambiguous due to part of speech. Some other instance of the same collocation in the same text may have become fully unambiguous. The section reports on an experiment with predicting the part-of-speech analysis of all instances of an ambiguous word sequence from the unambiguous analysis of at least one instance of that sequence.

  - More particularly, the section reports on an experiment with disambiguating with *noun groups* – a nominal head with at least one premodifier – as a case study. My hypothesis would in this case predict e.g. that the word-form token *head*, when preceded by the noun *cylinder*, will be a noun if the text contains another instance of the same word sequence fully disambiguated, and if the word sequence has been recognised as a noun group.

  - Analysis proceeds in the following fashion:

    1. The text is analysed with the grammar-based part of the ENGCG morphological disambiguator.

2. Unambiguous noun groups are extracted with an accurate noun group extraction program.

3. All (inflected and non-inflected) forms of the extracted word sequences are marked as noun groups in the original running text.

4. The ENGCG morphological disambiguator is applied to the same text again.

– An experiment was made with two texts totaling 215,000 words.

1. After the first round of disambiguation, 5.4% of all words remained ambiguous due to part of speech.

2. The noun group extractor produced a list of 16,100 distinct noun groups.

3. Applying the noun groups to the text made it possible to resolve almost 27% of those ambiguities that survived the ENGCG disambiguator during the first round.

4. The reliability of this mechanism was tested by proofreading a randomly selected sample of 2,000 word sequences that had been marked as noun groups in the text. No false hits were found. The mechanism seems to be extremely reliable, perhaps even more reliable than the grammar-based disambiguator.

– The reliability of the noun group mechanism suggests the question, how reliable the mechanism would be if noun groups extracted from one text were used in the analysis of another text, perhaps even from a different domain. If the text-generic application were reliable, a very large list of noun groups (and perhaps other kinds of collocations as well) could be generated from large corpora, and this kind of information would be a useful part of an ENGCG-style system.

## 2.3 *Designing a parsing grammar*

This study, published as a separate monograph, takes as its point of departure recent work within the Constraint Grammar framework. After a critical evaluation of Constraint Grammar and the English Constraint Grammar, we move on to Finite-State Intersection Grammar, a framework originally proposed by Koskenniemi [1990] and since then worked on also by Tapanainen, who has made operational finite-state parsers, and by the present author.

*INTRODUCTION*

- The concepts 'grammatical representation', 'recall' and 'precision' are introduced, and their relevance to the development and evaluation of parsing descriptions is shown.

- Recent work in parsing is examined in outline.

    - One approach is largely based on autonomous grammar theory. Emphasis has been on 'linguistically interesting' phenomena, and many properties of running text have been somewhat neglected. The grammatical representations are usually very detailed and ambitious, sometimes involving partly semantic distinctions. Parsing systems within this approach have usually been inaccurate in the analysis of running text: a large portion of input text is not recognised at all, and when a sentence is recognised, uncomfortably many analyses are usually proposed.

    - Probabilistic techniques have been used quite successfully especially at lower levels of structural analysis, and they have recently become something of a standard. A problem with this approach is the difficulty of improving the accuracy of the analysers beyond certain thresholds; for instance, to my knowledge the 97% accuracy has not been consistently exceeded with these techniques in part-of-speech annotation. Also, the application of these techniques to syntactic analysis has been somewhat tentative.

    - Various hybrid systems have also been proposed, for instance work jointly carried out by IBM and Lancaster University. Here the main idea is to let the linguist supply the description, and let statistical techniques take over the application of these rules in parsing. The results are promising.

    - 'Reductionistic surface-syntactic analysis' is a grammar-based approach pursued recently in Helsinki. Linguistic structure is coded with dependency-oriented functional tags. Parsing descriptions are produced by linguists rather than with statistical techniques. Morphology plays a central role; the lexicon is an inventory of morphosyntactic information. The parsing grammar is an unordered collection of grammar rules or constraints about the linear order of words and grammatical tags. Analysis proceeds as the sequence

        1. Context-free introduction of descriptors as alternatives
        2. Context-sensitive resolution of ambiguity

    The parser discards illegitimate alternatives; what survives the grammar is the analysis. Also heuristic routines can be used, though they have been somewhat marginal in our descriptions so far.

## CONSTRAINT GRAMMAR

A general introduction to and a critical evaluation of Constraint Grammar and the English parsing description is given.

- Constraint Grammar parsing is based on the sequential application of

  1. Preprocessing
  2. Introduction of morphological ambiguity: morphological analysis
     (a) TWOL-style lexical analysis
     (b) Analysis of unrecognised words: morphological heuristics
  3. Resolution of morphological ambiguity: morphological disambiguation
  4. Introduction of syntactic ambiguity: syntactic mapping
  5. Resolution of syntactic ambiguity: syntactic disambiguation

- The morphological disambiguator leaves the correct reading to 99.7–100% of all word-form tokens in running text, while 3–7% of all words remain partly ambiguous. The error rate of the syntactic module is 3–4%, and about 20% of all words remain syntactically ambiguous.

- An evaluation follows:

  - Merits:
    * The usefulness of the two-level model in the English morphological description only confirms previous experiences.
    * It is possible to write robust parsing descriptions in the Constraint Grammar framework, as indicated by the reliability of the disambiguation grammar. This robustness is made possible by the following factors:
      · The grammatical representation does not introduce certain unresolvable ambiguities.
      · The grammatical representation is well specified, i.e. the grammarian knows what he or she wants to express with the description.
      · Validating the grammar is easy because the consequence of each constraint is easily observable.
      · Our use of corpora has been extensive in the formulation and validation of the parsing descriptions.
      · Constraints can (and do) refer beyond the neighbouring word, up to sentence boundaries. This flexibility is not usually observable in probabilistic taggers.
      · Distributional generalisations can be expressed at various levels of abstraction, ranging from lexico-grammatical to more general syntactic categories. This contributes to the detail and exhaustivity of the grammar. An illustration is provided of how even uses less likely in terms of lexical probabilities can be uniquely and correctly identified, which is not usually the case with probabilistic systems.

- Pending problems that compromise the economy, transparency and accuracy of the description:

  * Identification of sentence-internal clause boundaries is not quite satisfactory.
  * A consequence of the sequentiality of the parsing architecture is that the grammatical representation at the non-final stages of analysis (e.g. during morphological disambiguation) is not sufficiently expressive.
  * The representation of ambiguity at the word level only detracts from the distinctiveness of the grammatical representation.
  * Reference to phrases and clauses is difficult.
  * Grammar rules are somewhat lacking in transparency because of their partiality and roundaboutness.

## FINITE-STATE INTERSECTION GRAMMAR

The latter half of the study reports on further developments in surface-syntactic analysis, this time within Finite-State Intersection Grammar, a framework proposed by Koskenniemi [1990]. First, a short introduction to this new formalism is given.

- As in Constraint Grammar, also here syntactic analysis operates on ambiguous sentence representations. However, here all types of descriptor (morphological, syntactic and word boundary tags) are introduced and processed in parallel. The input sentence is represented as a regular expression, and before parsing is translated into a finite-state automaton.

- Parsing means intersecting grammar automata with the sentence automaton. The intersection is the parse; the rest of the input sentence representation is what the grammar rejects.

- Ambiguity is represented at the sentence level (rather than the word level).

- The grammar is an unordered collection of rules.

- The rule formalism is flexible. The full power of extended regular expressions is available for representing sentences and rules. Both negative and positive distributional rules can be expressed. The 'implication' rule is the most useful: for a distributional category, all legitimate distributions can be listed as alternative context conditions in a readable and compact fashion.

- This framework overcomes certain problems that were observed in the Constraint Grammar formalism.

  - Clause boundaries are represented explicitly and in sufficient detail.

- All grammatical descriptors (i.e. all types of ambiguity) are represented in parallel.

- The rule formalism does not distinguish between morphological, clause boundary and syntactic tags, so all levels of ambiguity can be addressed simultaneously within a single rule component.

- Ambiguity is represented at the sentence level, which makes for the more complete resolution of ambiguity.

- It is easier to refer to phrases and clauses because regular expressions can be used in rules.

- The rule formalism contributes to the readability of the rules.

- An experiment with a 1,400-word text about morphological disambiguation is reported, using an experimental finite-state grammar, to test the previously stated assumptions that

  - Rule-based morphological disambiguation can be carried out further than was possible with ENGCG.

  - Morphological disambiguation can be carried out as a side effect of overtly syntactic rules (i.e. no separate grammar for morphological disambiguation is needed).

First, the text was analysed using the ENGCG morphological disambiguator. Then the syntactic representation, as specified in the experimental finite-state grammar, was introduced as new ambiguity. Then the finite-state parser was applied for resolving these syntactic ambiguities (and, as a side effect, the remaining morphological ambiguities). When syntactic and clause boundary information could be used, 40 out of the 43 ambiguities too hard for the ENGCG parser were resolved without errors.

The rules used in the experiment were mostly about syntactic functions. In other words, my results so far – the observation about the syntactic nature of the 23 statements underlying the 1,000 disambiguation constraints as well as this small-scale test – suggest the hypothesis that separate rules for morphological disambiguation are unnecessary, at least in the analysis of English.[8] Furthermore, it is suggested that if the functional description were extended to clauses, specific rules about clause boundaries would not be needed either.[9] Furthermore, the results suggest that part-of-speech disambiguation can be

---

[8]Due to the unfinished status of my current work within Finite-State Intersection Grammar, these observations should not be considered conclusive yet. The proof of this hypothesis presupposes a syntactic parsing grammar that is capable of carrying out also morphological disambiguation without using ENGCG as a 'preprocessor', so more conclusive evaluation of this hypothesis will be possible only at the maturation of the parsing grammar presently under development.

[9]As for the nature of those morphological ambiguities resolved by the ENGCG disambiguator, the rule typology proposed in *Morphological disambiguation* suggests that also here, the underlying

carried out further if direct reference to syntactic knowledge is possible in the grammar.

## A NEW GRAMMATICAL REPRESENTATION

This chapter in the monograph *Designing a parsing grammar* outlines a syntactic representation that is a redevelopment of the grammatical representations used in ENGCG and in my first experimental finite-state grammars. The main problem that faced these previous grammatical representations was a certain lack of expressiveness from the point of view of the grammarian. For example, the functional description was not extended to clauses. This made it difficult to express e.g. rules about the coordination of formally different function categories (for instance, the coordination subjects, one formally a noun phrase and the other a clause).

During the specification phase, the new representation was applied to the following corpora:

- About 2,000 sentences from *A Comprehensive Grammar of the English Language* by Quirk, Greenbaum, Leech and Svartvik [?]. The function of this grammar definition corpus is to represent the major syntactic constructions in English.

- Some 20,000 words of running text, which is tagged to represent at least the most frequent characteristics of authentic text that are not extensively treated in descriptive grammars. This corpus is being extended.

These tagged corpora can also be used effectively for testing the emerging parsing grammar.

The central characteristics of the new grammatical representation are the following:

- The tags indicate traditional syntactic functions in a dependency-oriented fashion.

- The description is underspecific with regard to certain phenomena that are not entirely resolvable on the basis of structural information.

- An explicit distinction is made between finite and nonfinite clauses, i.e. also clauses are furnished with a functional tag (Subject, Object etc.). In the less

---

generalisations are essentially syntactic. However, the proof of the reducibility of morphological disambiguation, and perhaps also of clause boundary determination, to syntax proper remains to be given at the maturation of the new syntactic finite-state description currently under development.

distinctive ENGCG grammatical representation, it is very difficult e.g. to employ the so-called Uniqueness Principle (maximally one potentially coordinated subject, object or other 'primary category' permitted in each clause) [Karlsson, 1985].

- The functional description is extended to finite and nonfinite clauses. This enables e.g. the statement of the distribution of clauses (as function categories).

- The new representation is economical. For instance, information that is encoded in the morphological tags is not repeated in the syntactic tags.

## GUIDELINES FOR THE GRAMMARIAN

This chapter shows how to write grammar rules in the Finite-State Intersection Grammar framework. First, a method is suggested, then sample rules from a parsing grammar are presented as an illustration.

- Formalising distributional statements.

  - The rule formalism is flexible; even new rule types can be defined if necessary. The *implication rule* appears to be the most expressive type of rule. The distribution of a grammatical category can be represented as alternative context conditions in an implication rule; the rule accepts the a reading containing the category only if the context is represented in the rule.
  - A starting point is to determine the distributional categories (along with their distributions). A comprehensive descriptive grammar is useful for this purpose. Also testing the rules against running text may reveal further generalisations.
  - The next question is how to represent the constructions as regular expressions, e.g. how to represent a declarative clause containing a subject, or a direct question containing a subject, or a tag question containing a subject, and so on.
  - An example is provided, showing how underspecificity can be employed to bring out only the linguistically relevant properties of the generalisation. This contributes to the readability and compactness of the grammar.
  - The constructions can then be listed as possible contexts for the distributional category under discussion, which itself is given to the left of the implication arrow.

- Finally, substantial examples from an emerging grammar of 70 implication rules are given from the description of simple noun phrases and prepositional phrases to suggest that a realistic parsing grammar can also account for less frequent, or 'marginal', constructions.

*ON HEURISTICS*

This final chapter presents a mechanism for using heuristics that can be used to choose among alternative analyses. A few applications are outlined:

- Heuristics for preferring frequent function assignments.

- Heuristics for the recognition of collocations, e.g. multi-word compounds.

- Heuristics for the recognition of typical head–argument structures.

- A mechanism for employing lexical probabilities.

# Chapter 3

# Orientation

The three papers that constitute the bulk of this dissertation mainly address various aspects of the general question, how to write an accurate parsing grammar. In part, my answer to the general question is given as documentation of existing descriptions to which I have contributed; in part, as the design of future descriptions.

Relatively little is told about the significance of the present work to the field at large; this is especially true about *Morphological disambiguation* and *Experiments with heuristics*, both chapters in a book where most of the general orientation is provided by Karlsson [1994]. The separately published monograph *Designing a parsing grammar* is a little more self-contained; some perspective to other work is sketched in the introductory chapter, and also considerable attention is devoted to relating two brands of reductionistic surface-syntactic analysis, namely Constraint Grammar and Finite-state Intersection Grammar, to each other.

Let us outline the research goal in a little more detail. The point of departure is the thesis that it is possible to make a competitive rule-based parsing description of a natural language, for the purposes of analysing (relatively) unconstrained running text at the level of part-of-speech analysis. In particular, we are talking about manually made descriptions based on linguistically motivated generalisations rather than about (rule-based) systems that are (for the most part) generated automatically from corpora (for instance, cf. [Hindle, 1989]).

*Morphological disambiguation* is mainly concerned with this thesis. The problem addressed in this paper is rule-based morphological or part-of-speech disambiguation of English. The problem is well chosen because there are more than ten taggers with an accuracy of 95–97%, each based on statistical methods.[1] To the best of the author's knowledge, there were no competitive rule-based part-of-speech taggers in existence when work on the ENGCG description was started.

---

[1]Interestingly, this performance level was first reached over ten years ago with the CLAWS1 tagging system [Marshall, 1983], and no significant improvement has been achieved since then within this approach.

The spectacular success of statistical methods, as compared to the relatively modest performance of earlier rule-based systems in this problem area (cf. [Greene and Rubin, 1971]), serves as a challenging benchmark for any rule-based system. Proposing a rule-based system with equal or better accuracy would certainly be news worth telling – not only because of the relatively long life of the reported 'world record' accuracy of 95–97%, but also because of certain arguments voiced against even the possible viability of the rule-based approach in this area (cf. [Sampson, 1987a; Church, 1992]). Thus success with a grammar-based approach in this area would also contribute to the methodological controversy between the rule-based and statistical approaches.

The chapter *Morphological disambiguation* documents and evaluates a grammar for part-of-speech disambiguation that meets this challenge, at least in a particular sense: if a little ambiguity is tolerated in the tagger's outputs, the error rate of our grammar-based ENGCG tagger seems to be about ten times smaller than that of other state-of-the-art systems under similar conditions (i.e. with the same amount of ambiguity in the output). In other words, *Morphological disambiguation* serves as a practical proof that it is possible to write a competitive parsing description for this problem area; in this way, the chapter also contributes to the methodological debate.

The focus of the other two articles in this dissertation shifts somewhat, from proving the competitiveness of the rule-based approach to the design of even more accurate and linguistically better motivated parsing grammars. The chapter *Experiments with heuristics* contributes to this by showing how the rule-based approach can also be used for employing tendencies rather than absolute regularities in resolving the remaining ambiguities. Three promising techniques are outlined for employing lexical and syntactic information.

The third study, *Designing a parsing grammar*, takes a more radical departure from the ENGCG work. Here we are again concerned with the rule-based approach, but at a higher ambition level than was the case with my previous work. Already in *Morphological disambiguation*, I argued for a more uniform grammatical representation by showing how the sequentiality and unoptimal interaction between the various grammar components detracted from the accuracy of the resulting parser. In *Designing a parsing grammar*, this idea is taken up again. It is argued that it is possible to dispense with modularity within the grammar by using a parallel rather than sequential grammatical representation, and by using a more uniform and powerful rule formalism – both available within the Finite-State Intersection Grammar framework originally proposed by Koskenniemi [1990] and later on refined by Tapanainen and the present author [Tapanainen, 1991; Koskenniemi, Tapanainen and Voutilainen, 1992; Voutilainen and Tapanainen, 1993].

More particularly, it is argued that what was represented as three different grammar components in the Constraint Grammar framework can be represented as a monolithic, descriptively uniform rule component: clause boundary determination, morphological (or part-of-speech) disambiguation and proper syntactic analysis seem to be reducible to a single grammar that overtly is about the form and function of everyday syntactic categories such as phrase-like units, verb chains, modifier–head relations, and syntactic functions (subjects, objects and so on). In support of this hypothesis, also a small-scale experiment is reported.[2]

Achieving a uniform, syntax-oriented and robust parsing description presupposes a carefully specified, structurally resolvable and sufficiently expressive grammatical representation. *Designing a parsing grammar* outlines a new representation that is derived from the ENGCG syntactic representation but is more expressive with regard to the representation of phrases and clauses, thus enabling the grammarian to state structural generalisations in a more transparent and exhaustive manner than was possible in ENGCG. A new parsing grammar based on this new representation is currently under development.

Next, the following topics will be discussed in more detail:

- On the possibility of specifying the grammatical representation

- Two approaches to structural computer analysis of running text

- Combining grammar-based and probabilistic techniques

- Designing a functional dependency-oriented grammatical representation

## 3.1 On the possibility of specifying the grammatical representation

The design and evaluation of a parsing description presupposes an answer to the question what is the appropriate analysis of any utterance in the object language in terms of the employed grammatical representation – the convention according to which grammatical descriptions are assigned. In plain English: before making the parsing description, the grammarian should know, what she or he wants to say in the first place. Whenever an answer to this question cannot be given, the design or evaluation of the description with regard to the question remains unsatisfactory.

Is it possible to reach a near-100% consensus about the appropriate analysis? The issue appears to be somewhat controversial. Sampson [1987b; forthcoming] seems to admit the possibility. This is also manifest in the syntactically annotated SUSANNE corpus, annotated by Sampson and his colleagues. Also recent annotation work on the creation of a 800,000-word skeleton parse bank, carried out at Lancaster University [Garside and McEnery, 1993; Eyes and Leech, 1993] seems to support the possibility of near-100% consensus.[3]

---

[2]However, due to the unfinished status of my work in this new framework, no final conclusions can be drawn as yet.

A much more pessimistic estimate is proposed by Church [1992]: he states that linguistic experts part-of-speech annotating the same corpus will agree only in 95% of all cases, and the majority of the 5% residual represents a problem where a consensus cannot be reached even after considerable negotiations. This would imply that claiming a recall of close to 100% in part-of-speech analysis – as I do regarding the ENGCG morphological disambiguator – could not be taken seriously, because, by definition, claiming a near-100% recall presupposes at least a near-100% consensus about the correct analysis.

Our experiences from the ENGCG description are very much in line with [Eyes and Leech, 1993], and so in a sharp conflict with [Church, 1992]. Two experiments on manual part-of-speech analysis are reported in *Morphological disambiguation* (Section 1.4): first, three people familiar with the ENGTWOL grammatical representation independently performed the role of the disambiguator, marking the contextually most appropriate alternative in all ambiguous ENGTWOL analyses in a 500-word text; second, the three judges manually assigned the ENGTWOL part-of-speech categories to another 500-word text, this time performing the combined role of the morphological analyser and the disambiguator. Both experiments gave similar results. After a first, automatic comparison, the analyses of any two judges agreed more than 99% of the time, and after the differences were examined collectively, they were agreed to be errors due to inattention; that is, no genuine differences in opinion came up.

These experiments, along with our previous experiences, suggest that with a carefully specified grammatical representation, a near-100% consensus can be reached, at least at a relatively low level of analysis.[4]

---

[3]Eyes and Leech [1993] mention a correctness rate of 99.9% in part-of-speech tagging after human postediting; a sensible interpretation of this figure implies that a consensus is reached about the appropriate analysis in at least 99.9% of all cases.

[4]However, very categorical conclusions about reaching a 100% consensus should be avoided here because of the size of the data used in the experiments. More extensive experiments are needed for getting a stable view of a realistic agreement rate. Nevertheless, the data appears large enough for refuting the very pessimistic estimates given in [Church, 1992]. – It also remains to be investigated, how much disagreement will come up at higher levels of analysis; our experience is that at least the ENGCG syntactic analysis is not very different from part-of-speech disambiguation in this respect, though probably the effort of specifying a grammatical representation becomes more laborious the more delicate the proposed grammatical representation is.

## 3.2  Two approaches to structural computer analysis of running text

There are two main approaches to the structural analysis of running text, the *grammar-based* and the *probabilistic*.[5] Typically, the parsing description in a grammar-based system is a system of linguistic hand-written rules, while a probabilistic analyser normally employs an automatically acquired corpus-based statistical database that may linguistically appear quite unrevealing.

Several grammar-based and probabilistic parsing systems have been proposed in recent years for different types of structural analysis, e.g. part-of-speech tagging and syntactic analysis.

Generally, both probabilistic and rule-based systems have had a rather modest success in the full-scale syntactic analysis of running text (for instance, see discussions in [Black, 1993; Karlsson, 1994]). In part-of-speech tagging of running text, on the other hand, probabilistic systems have been quite successful, often reaching a 95–97% accuracy. For the analysis of English alone, more than ten probabilistic taggers with this accuracy have been proposed during the last ten years (see bibliography in [Church, 1992]). To my knowledge, no serious grammar-based competitors were proposed before the introduction of the English Constraint Grammar parser [Karlsson, Voutilainen, Heikkilä and Anttila (eds.), 1994]. Perhaps in the wake of the success of the probabilistic approach in low-level structural analysis, some more general methodological arguments against the viability of the grammar-based approach have also been proposed e.g. to the effect that

1. Accounting exhaustively for the variety of a natural language in a grammar is likely to fail for the purposes of analysing running text; therefore the grammar-based approach will be inferior [Sampson, 1987a].

2. Running text contains ill-formed and deviant utterances; such utterances are by definition not accounted for by a grammar that makes a distinction between the acceptable and the ill-formed; therefore a grammar-based system will fail to analyse ill-formed utterances; therefore the grammar-based approach is likely to be impracticable [Sampson, 1987a].

3. Lexical likelihoods are more reliable that contextual likelihoods; therefore context contributes less (to part-of-speech disambiguation) than lexical likelihoods. Grammar-based systems prefer context and ignore lexical likelihoods (in part-of-speech disambiguation); therefore the grammar-based approach will be less successful (adapted from [Church, 1988; Church, 1992]).

---

[5]These are by no means the only ones, though probably most systems employ techniques from either or both of these two.

Let us examine these arguments. Sampson may be right in claiming that an exhaustive parsing description is not likely to emerge. This incompleteness of the description, however, does not entail the inferiority of the grammar-based approach because there are no good reasons for believing that perfection can be reached with any other approach either. Certainly, e.g. a probabilistic tagger will make some prediction about almost any input, but, because of the inherent proneness for errors of the approach, this kind of predictiveness hardly implies that a probabilistic system is exhaustive in the sense that the grammar-based system may ever fail to be.

Regarding Sampson's second argument, I agree with his point that a grammar will not account for ill-formed utterances. However, it is a different thing to claim that a grammar-based *system* will fail to analyse ill-formed input. I point out in *Morphological disambiguation* that the application of the grammar itself can be constrained if necessary. For instance, a reductionistic parser can 'refuse' to consult the grammar in the case of unambiguous input. Also, it is shown in *Experiments with heuristics* as well as in *Designing a parsing grammar* that a grammar-based system can employ rules based on linguistic generalisations, though somewhat rougher ones than some others, and these 'heuristic' rules can be applied in a less categorical manner.

Regarding Church's claim, I again do not object to his observation that lexical likelihoods are more reliable than contextual ones. The flaw in his argument is the equation of contextual likelihoods with context in general. The information-theoretic method of employing context is only one possibility, perhaps a poor one; in any case it does not follow that *all* ways of using context will fail because one does. This issue is examined more closely in *Designing a parsing grammar*.

The performance of the ENGCG morphological disambiguator itself is of course a compelling argument for the viability of the grammar-based approach. Throughout the dissertation, the solidity of the grammar-based approach is shown from various points of view; here, three illustrations are given:

- The ENGCG morphological disambiguator does not resolve all ambiguities, but what it does, it does very reliably. In *Morphological disambiguation* it is shown that with 1.04 morphological readings per word in the disambiguator's output, ENGCG had a recall of 99.8%, i.e., only 0.2% of all words missed the correct morphological analysis. These figures are compared to another probabilistic state-of-the-art tagger [de Marcken, 1990], and the comparison suggests that with the same amount of ambiguity in the output, the probabilistic system misses a correct analysis about ten times more often than ENGCG does. (Further empirical evidence in this direction is given e.g. in [Voutilainen and Heikkilä, 1994b].)

- All predictions made by the morphological disambiguator are entirely based on context. After the combined effect of the grammar-based and heuristic ENGCG constraints, 96% or more of all words in the analysed text become correctly and uniquely identified.[6] Compare this to the near-90% accuracy achieved with lexical probabilities alone [Church, 1992]: contrary to Church's claim, context, at least as used in ENGCG, contributes more to part-of-speech disambiguation than lexical probabilities do, by some six percentage points at least.

- Regarding the resolution of those ambiguities that survive the present ENGCG disambiguator, it is argued that most of even these ambiguities are structurally resolvable; the main reason why they are not resolved by the ENGCG disambiguator is the shortcomings in the design of the system and of the ENGCG grammatical representation. For instance, a small-scale experiment reported in *Designing a parsing grammar* gives empirical support to this idea.[7]

All in all, combining a flexible, expressive rule formalism with an empirical approach to writing the parsing description makes for a very dependable and robust analyser. I do not claim that lexical factors should be ignored altogether; only, so far I have been interested in seeing how far a syntax-oriented description can take us. As suggested in *Experiments with heuristics*, using lexical knowledge, especially collocational information, is likely to be a very reliable add-on to syntax-oriented rules. Probably syntactic and lexical information together, used in a regimented fashion, make for a more accurate analyser than either separately.

## 3.3 Combining grammar-based and probabilistic techniques

Traditionally, probabilistic systems acquire their language models from one, perhaps very heterogeneous corpus like the tagged Brown University Corpus or its

---

[6]The remaining four per cent are ambiguous, but for the most part also these retain the correct analysis; no more than 0.5% of all words in the ENGCG tagger's output miss the correct analysis.

[7]Also my most recent work within the Finite-State Intersection Grammar framework bears on this issue. [Voutilainen, 1994] reports on an experiment with two different texts (5,100 words in all) where the output of the grammar-based part of the ENGCG morphological disambiguator is enriched with the syntactic representation outlined in *Designing a parsing grammar*, and then the new ambiguity is resolved using the latest version of the new finite-state grammar that I am currently writing. When the finite-state parser was forced to produce exactly one parse per sentence, the output became fully unambiguous, also with regard to part of speech, with an overall accuracy of 99.5% at the part-of-speech level, as defined in the ENGTWOL description (i.e. only 0.5% of all words missed the correct part-of-speech analysis in the output). This accuracy, achieved using nothing but grammar rules for disambiguation, compares quite favourably with the 95–97% accuracy achieved with other taggers so far.

Also compare this context-based accuracy of 99.5% with the near-90% accuracy achieved using lexical probabilitites alone; these figures seem to suggest, in contrast with [Church, 1992], that context is considerably more important for part-of-speech disambiguation than lexical probabilities are.

British counterpart, the tagged Lancaster–Oslo/Bergen Corpus, and use this language model in the analysis of new texts. If the analysed text happens to differ significantly (in terms of the language model) from the corpus the language model was based on, many predictions of the analyser are likely to go amiss.

In *Experiments with heuristics*, I explore the possibility of acquiring (some parts of) the language model from the text to be analysed itself. The idea is to derive the language model from that part of the text which has been fully disambiguated by the very dependable grammar-based system, e.g. the ENGCG disambiguator, and use this model in the analysis of those ambiguities that the grammar-based system left pending.

Two empirical studies are documented, one on acquiring and using certain kinds of lexical probabilities, the other on collocation-based disambiguation (learning modifier–head sequences). The results are quite promising, supporting the apparently self-evident idea that a text represents itself better than some other text does.

This kind of combination could benefit from the best sides of grammar-based and statistical techniques. On the one hand, the grammar-based analyser would produce a very reliably tagged 'learning corpus' of the main part of the original text, in this way enabling the acquisition of the *relevant* language model by the statistical system. On the other hand, the statistical system, whose language model is based on the relevant material, may be able to predict so accurately about the *structurally* hard cases that the grammar-based system's unability to resolve all ambiguity would not be so problematic from a practical point of view.[8]

## 3.4 Designing a functional dependency-oriented grammatical representation

The rather modest success of parsers based on autonomous grammar theory on parsing running text (see e.g. [Black, 1993; Karlsson, 1994]) is in part due to the design of the grammatical representations that these parsers employ. As observed in *Designing a parsing grammar*, many of the distinctions made in these grammatical representations are not entirely structural in the sense that also semantic and higher-level knowledge is needed for the knowledge-based accurate employment of these distinctions in a parsing system. If such higher-level knowledge is not available, the options are either to leave the ambiguity unresolved (which often implies a multitude of alternative parses) or to risk a guess (which tends to result in the loss of the most appropriate analysis).

---

[8]Nor would it be necessary for the statistical system to resolve all those ambiguities left pending after the first application of the grammar-based system; even the resolution of some of the remaining ambiguities might suffice because the grammar-based system might be able to carry on, as a domino effect. Perhaps the grammar-based and statistical modules could best operate in parallel, each resolving an ambiguity if a common, very high reliability threshold is exceeded, while the learning component of the latter module would be active throughout the analysis, learning from previous predictions. In this way, both modules could feed each other and so carry on with the analysis, with a very small risk of misanalysis.

In the design of a grammatical representation for use in a rule-based parsing description, the introduction of structurally unresolvable distinctions should be avoided. It seems preferable to assign a shallow but accurate analysis rather than an ambitious but unreliable analysis.

To a large extent, this desideratum of *shallowness* is already met in the ENGCG grammatical representation as well as its predecessor, the FPARSE representation [Karlsson, 1985]. For instance, many unresolvable ambiguities due to the attachment of adverbials and due to modifier scope are never introduced. However, working with a shallow grammatical representation brought up some new observations. Firstly, when I wrote the grammar for morphological disambiguation in the Constraint Grammar framework, it became obvious that much of the part-of-speech ambiguity remains unresolved because the available grammatical representation was not particularly *expressive*. It is argued in *Morphological disambiguation* that the generalisations underlying the constraints are actually partial and very roundabout paraphrases of the form and function of syntactic categories. Some empirical evidence is also given in *Designing a parsing grammar*, where the hypothesis that part-of-speech disambiguation can be carried out as a side effect of proper syntactic rules was tested with regard to those part-of-speech ambiguities that the ENGCG morphological disambiguator was unable to resolve. In short: what the disambiguation grammarian found himself talking about was more general syntactic statements, but, because of the problems in the expressiveness of the available grammatical representation (morphological tags only), these essentially syntactic generalisations had to be expressed in a roundabout and partial fashion.

These first observations suggest that the expressiveness of the grammatical representation could be much improved if all types of descriptor, ranging from morphological to syntactic function tags, were accessible simultaneously to a single uniform rule component. No modularity in the form of separate, sequentially applied subgrammars, should be enforced; the appropriate manner of expressing the relevant generalisations should be left to the discretion of the grammarian.

In my experimental descriptions within the Finite-State Intersection Grammar formalism, reported in *Designing a parsing grammar* (see also [Koskenniemi, Tapaninen and Voutilainen, 1992; Voutilainen and Tapanainen, 1993]), all three types of descriptor – morphological, clause boundary, and syntactic tags – were resolved in parallel. The grammar consisted of a single rule component. The formalism made no distinction between these three types of descriptor, so the grammarian was free to address any type of ambiguity with any kind of rule.

My experiments with part-of-speech disambiguation using overtly syntactic rules suggested the hypothesis that all three types of ambiguity could almost entirely be resolved with rules that overtly concerned the form and function of syntactic cate-gories. However, in addition to the predominating syntactic-functional rules, there were also some rules overtly about clause boundary tags. On closer inspection, it appears that also these few clause boundary rules are roundabout expressions of the form and function of syntactic categories – in particular, about the distribution of clauses as functional categories (clauses as postmodifiers, subjects, objects, and so on). Again, the roundaboutness in the grammar seems to be due to the short-comings in the expressiveness of the grammatical representation. At that point of development, the representation already made an explicit distinction between finite and nonfinite clauses, but the functional account had not been extended to the description of clauses. If the grammarian wanted to express a generalisation about clauses as a functional category, it had to be done in a roundabout, and probably incomplete fashion – much the same type of situation that was so predominant in earlier ENGCG work.

In *Designing a parsing grammar* (see also [Voutilainen and Tapanainen, 1993]) I propose a new version of a functional dependency-oriented grammatical representation. Let us recapitulate the main properties of this representation.

- It is derived from the ENGCG representation.

- It expresses syntactic functions in an underspecific dependency-oriented fashion. Each word is provided with at least one tag that indicates its syntactic function (e.g. Subject, Premodifier, Auxiliary, Preposition complement).

- Boundaries of finite clauses are indicated with clause boundary tags. An explicit distinction is made between boundaries between juxtaposed full clauses and boundaries separating a centre-embedded clause.

- The functional description is extended to clauses. The function of each clause (finite or nonfinite) is indicated by assigning a clause function tag to each main verb. The functional account of clauses has several desirable consequences for the grammarian, for example:

  - The distribution of clauses can be described more easily because many distributional restrictions concern functional rather than form categories. Expressing distributional restrictions on clauses (as functional categories) has as a desirable side effect the elimination of contextually illegitimate clause boundaries. In other words, specific clause boundary rules may turn out to be unnecessary when a sufficiently expressive grammatical representation is available.

  - The coordination of formally different function categories becomes easier to describe, witness e.g. the coordination of a nominal subject and a *WH*-clause acting as a subject.

– From the function of a clause, it is possible to infer whether the clause is a main clause or a subordinate clause. Because word (or rather constituent) order is more fixed in a subordinate clause, knowing the clause type makes it possible to express more severe word order restrictions within a subordinate clause.

- An explicit distinction is made between finite and nonfinite clauses. Consequently,

  – The employment of the so-called 'Uniqueness Principle'[9] [Karlsson, 1985] becomes easier to employ. Without the distinction, one finite clause could contain nonfinite clauses, and if e.g. subjects in nonfinite clauses could not be distinguished from subjects in finite clauses, accounting for both subject types at once would be very difficult. With the distinction between finite and nonfinite clauses, the Uniqueness Principle can be imposed on finite clauses without worrying about nonfinite ones, and vice versa. – In FPARSE [Karlsson, 1985], numerical indexing was used. With the present grammatical representation, it is possible to dispense with this indexing mechanism. In a manner of speaking, form categories can here be 'indexed' to function categories directly.

  – Word order is more fixed in nonfinite clauses than in finite clauses. Word order restrictions can be imposed more effectively if the clause type is known, as is the case here.

- From the parsing point of view, one could make the observation that with this new grammatical representation, the input will contain much more ambiguity than would be the case with a more ascetic grammatical representation. One could now argue that because of this massive ambiguity, parsing will become too slow to be practicable. However, our experiences (see [Voutilainen and Tapanainen, 1993]) suggest that parsing speed is determined by the resolvability of the ambiguity rather than by the amount of ambiguity in the input as such.

---

[9]'Each clause contains no more than one (potentially coordinated) primary function (e.g. Subject and Object).'